\documentclass{article}

\usepackage[nonatbib,final]{nips_2016}

\usepackage{listings}
\usepackage{graphicx}
\usepackage{color}
\usepackage{hyperref}

\usepackage{listings}
\usepackage[utf8]{inputenc} 
\usepackage[T1]{fontenc}    
\usepackage{hyperref}       
\usepackage{url}            
\usepackage{booktabs}       
\usepackage{amsfonts}       
\usepackage{nicefrac}       
\usepackage{microtype}      
\usepackage{amsmath}        

\title{Towards the Automatic Anime Characters Creation with Generative Adversarial Networks}

\author{
  Yanghua Jin \\
  School of Computer Science \\
  Fudan University\\
  \texttt{jinyh13@fudan.edu.cn} \\
   \And
   Jiakai Zhang \\
   School of Computer Science \\
   Carnegie Mellon University \\
   \texttt{jiakaiz1@andrew.cmu.edu} \\
  \AND
  Minjun Li \\
  School of Computer Science \\
  Fudan Univerisity \\
  \texttt{minjunli13@fudan.edu.cn} \\
  \And
  Yingtao Tian \\
  Department of Computer Science \\
  Stony Brook University \\
  \texttt{yittian@cs.stonybrook.edu} \\
  \And
  Huachun Zhu \\
  School of Mathematics \\
  Fudan Univerisity \\
  \texttt{zhuhc14@fudan.edu.cn} \\
  \And
  Zhihao Fang \\
  Department of Architecture \\
  Tongji Univerisity \\
  \texttt{fangzhihao126@gmail.com}
}

\begin{document}

\maketitle

\begin{abstract}
Automatic generation of facial images has been well studied after the Generative Adversarial Network(GAN) came out. There exists some attempts applying the GAN model to the problem of generating facial images of anime characters, but none of the existing work gives a promising result. In this work, we explore the training of GAN models specialized on an anime facial image dataset. We address the issue from both the data and the model aspect,
by collecting a more clean, well-suited dataset and leverage proper, empirical application of DRAGAN.
With quantitative analysis and case studies we demonstrate that our efforts lead to a stable and high-quality model.
Moreover, to assist people with anime character design, we build a website\footnote{\url{http://make.girls.moe}} with our pre-trained model available online, which makes the model easily accessible to general public.

\end{abstract}
\section{Introduction}


We all love anime characters and are tempted to create our custom ones.
However, it takes tremendous efforts to master the skill of drawing, after which we are first capable of designing our own characters.
To bridge this gap, the automatic generation of anime characters offers an opportunity to bring a custom character into existence without professional skill.
Besides the benefits for a non-specialist,
a professional creator may take advantages of the automatic generation for inspiration on animation and game character design;
a Doujin RPG developer may employ copyright-free facial images to reduce designing costs in game production. 

Existing literature provides several attempts for  
generation facial images of anime characters. 
Among them are Mattya\cite{ChainerDCGAN} and Rezoolab\cite{RezoolabDCGAN}
who first explored the generation of anime character faces right after the appearance of DCGAN\cite{DCGAN}. 
Later, Hiroshiba\cite{GirlFriendFactory} proposed the conditional generation model for anime character faces. 
Also, codes are made available online focusing on anime faces generation such as IllustrationGAN\cite{IllustrationGAN} and AnimeGAN\cite{AnimeGAN}. 
However, since results from these works are blurred and distorted on an untrivial frequency,
it still remains a challenge to generate industry-standard facial images for anime characters.

In this report, we propose a model that produces anime faces at high quality with promising rate of success. 
Our contribution can be described as three-fold:
A clean dataset, which we collected from Getchu, a suitable GAN model, based on DRAGAN,
and our approach to train a GAN from images without tags, which can be leveraged as a general approach to training supervised or conditional model without tag data.

\section{Related Works}

Generative Adversarial Network (GAN) \cite{GAN}, proposed by Goodfellow et al., shows impressive results in image generation \cite{DCGAN}, image transfer\cite{Img2Img}, super-resolution\cite{ledig2016photo} and many other generation tasks.
The essence of GAN can be summarized as training a \textit{generator} model and a \textit{discriminator} model simultaneously, where the discriminator model tries to distinguish the real example, sampled from ground-truth images, from the samples generated by the generator. 
On the other hand, the generator tries to produce realistic samples that the discriminator is unable to distinguish from the ground-truth samples.
Above idea can be described as an \textit{adversarial loss} that applied to both generator and discriminator in the actual training process, which effectively encourages outputs of the generator to be similar to the original data distribution.


Although the training process is quiet simple, optimizing such models often lead to \textit{mode collapse}, in which the generator will always produce the same image. 
To train GANs stably, Metz et al. \cite{metz2016unrolled} suggests rendering Discriminator omniscient whenever necessary. 
By learning a loss function to separate generated samples from their real examples, LS-GAN\cite{qi2017loss} focuses on improving poor generation result and thus avoids mode collapse.
More detailed discussion on the difficulty in training GAN will be in Section \ref{improved_gan_training}.

Many variants of GAN have been proposed for generating images. 
Radford et al. \cite{DCGAN} applied convolutional neural network in GAN to generate images from latent vector inputs.
Instead of generating images from latent vectors, serval methods use the same adversarial idea for generating images with more meaningful input. 
Mirza \& Osindero et al. introduced Conditional Generative Adversarial Nets \cite{cGAN} using the image class label as a conditional input to generate MNIST numbers in particular class.
Reed et al. \cite{reed2016generative} further employed encoded text as input to produce images that match the text description. 
Instead of only feeding conditional information as the input, Odena et al. proposed ACGAN\cite{odena2016conditional}, which also train the discriminator as an auxiliary classifier to predict the condition input.

\section{Image Data Preparation}
It is well understood that image dataset in high quality is essential, if not most important, to the success of image generation.
Web services hosting images such as Danbooru\footnote{\url{danbooru.donmai.us}} and Safebooru\footnote{\url{safebooru.org}}, 
commonly known as image boards,
provide access to a large number of images enough for training image generation models.
Previous works mentioned above all base their approaches on images crawled from one of these image boards,
but their datasets suffer from high inter-image variance and noise.
We hypothesize that it is due to the fact that image boards allow uploading of images highly different in style, domain, and quality,
and believe that it is responsible for a non-trivial portion of quality gaps between the generation of real people faces and anime character faces.
In order to bridge such a gap, 
we propose to use a more consisting, clean, high-quality dataset,
and in this section we introduce our method of building such a dataset.

\subsection{Image Collection}
\begin{figure}
  \centering
  \includegraphics[width=5in]{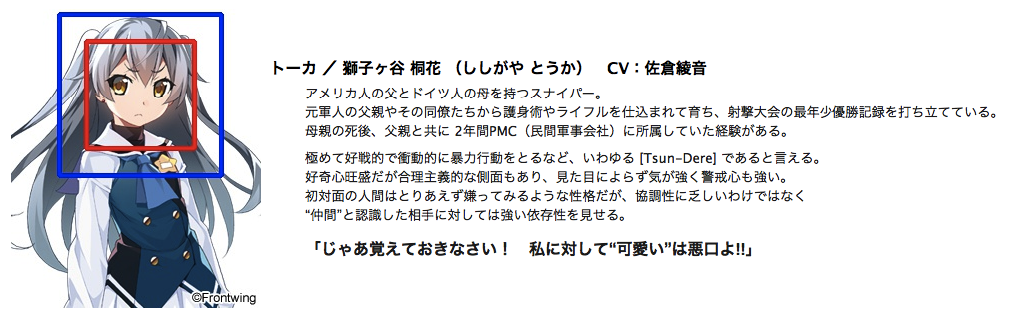}
  \caption{Sample Getchu page and the detection result,\url{http://www.getchu.com/soft.phtml?id=933144}.  Red line indicate the original bounding box and blue line indicate the scaled bounding box. Copyright: Frontwing, 2017}
  \label{fig:getchu_sample}
\end{figure}

Getchu \footnote{\url{www.getchu.com}} is a website providing information and selling of Japanese games,
for which there are character introduction sections with
standing pictures.
Figure \ref{fig:getchu_sample} shows one sample character introduction from the site. 
These images are diverse enough since they are
created by illustrators with different styles for games in a diverse sets of theme, yet consisting since they are all belonging to domain of character images, are in descent quality, and are properly clipped/aligned due to the nature of illustration purpose.
Because of these properties, they are suitable for our task.

Our collection of images consists of the following steps.
First we execute a SQL query on ErogameScape's Web SQL API page\footnote{\url{http://erogamescape.dyndns.org/~ap2/ero/toukei_kaiseki/sql_for_erogamer_form.php}} to get the Getchu page link for each game. The SQL query we used is described in Appendix \ref{SQL code}.
Then we download images and apply lbpcascade animeface \footnote{\url{https://github.com/nagadomi/lbpcascade_animeface}},
an anime character face detector, 
to each image and get bounding box for faces. 
We observe that the default estimated bounding box is too close to the face to capture complete character information that includes hair length and hair style, so we zoom out the bounding box by a rate of 1.5x. The difference is shown in Figure \ref{fig:getchu_sample}. 
Finally, from 42000 face images in total from the face detector,
we manually check all anime face images and remove about $4\%$ false positive and undesired images.

\subsection{Tag Estimation}
The task of generating images with customization requires categorical metadata of images as priors. 
Images crawled from image boards are accompanied by user created tags which can serve as such priors, as shown in previous works.
However, Getchu does not provide such metadata about their images,
so to overcome this limitation, we propose to use a pre-trained model for (noisy) estimations of tags. 

We use Illustration2Vec\cite{saito2015illustration2vec}, a CNN-based tool for estimating tags of anime illustrations\footnote{Pre-trained model available on \url{http://illustration2vec.net/}} for our purpose.
Given an anime image, this network can predict probabilities of belonging to
512 kinds of general attributes (tags) such as ``smile'' and ``weapon'',
among which we select 34 related tags suitable for our task. 
We show the selected tags and the number of dataset images corresponded to each estimated tag in Table \ref{tab:tag_count}.
For set of tags with mutual exclusivity (e.g. hair color, eye color), 
we choose the one with maximum probability from the network as the estimated tag. 
For orthogonal tags (e.g. ``smile'', ``open mouth'', ``blush''), 
we use $0.25$ as the threshold and estimate each attribute's presence independently. 

\begin{table}
  \begin{tabular}{c c c c c c c}
    \toprule
    blonde hair & brown hair & black hair & blue hair & pink hair & purple hair & green hair \\ 
    4991 &   6659&   4842 &   3289 &   2486 &   2972&    1115  \\
    \midrule
    red hair & silver hair & white hair & orange hair & aqua hair & gray hair & long hair \\  
    2417 &    987&    573&    699&    168&     57&  16562
  \\   
    \midrule 
    short hair & twintails & drill hair & ponytail & blush & smile & open mouth \\ 
    1403 &   5360&   1683&   8861&   4926&   5583&   4192 \\
    \midrule
    hat & ribbon & glasses & blue eyes & red eyes & brown eyes & green eyes \\ 
     1403 &   5360 &  1683&   8861&   4926&   5583&   4192 \\
    \midrule
    purple eyes & yellow eyes & pink eyes & aqua eyes & black eyes & orange eyes \\ 
    4442 &   1700 &    319 &    193&    990&    49 \\ 
    \bottomrule
  \end{tabular}
  \caption{Number of dataset images for each tag}
  \label{tab:tag_count}
\end{table}

\begin{figure}
  \centering
  \includegraphics[width=6in]{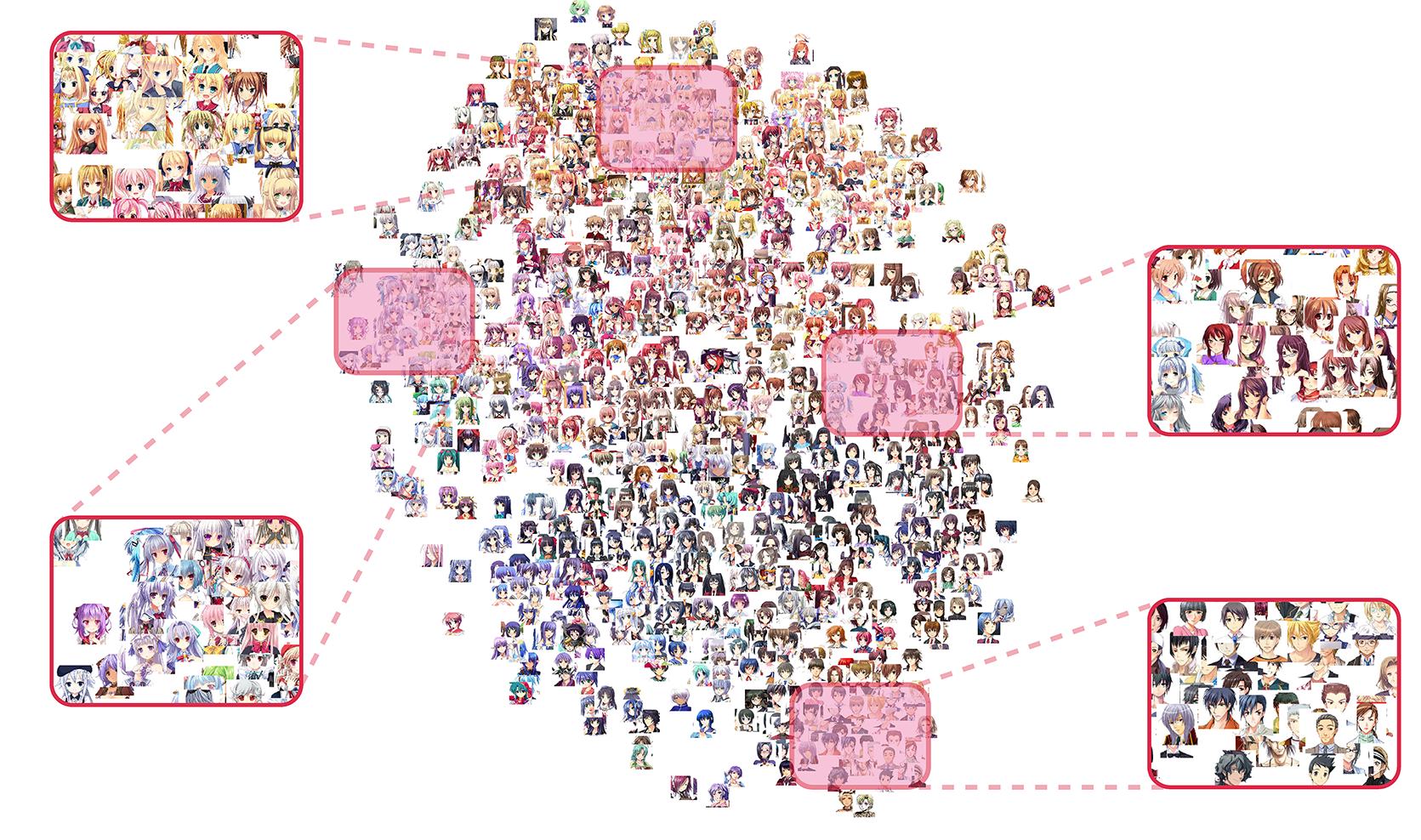}
  \caption{t-SNE visualization of 1500 dataset images}
  \label{fig:t_sne}
\end{figure}

\subsection{Visualization}

We would like to show the image preparation and the performance of tag estimation through visualization.
As an approximation, we apply the Illustration2Vec feature extractor,
which largely shares architecture and weights with Illustration2Vec tag estimator,
on each image for a 4096-dimension feature vector, 
and project feature vectors onto a 2D space using t-SNE\cite{maaten2008visualizing}.
Figure \ref{fig:t_sne} shows the t-SNE result of $1500$ images sampled from the dataset. We observe that character images with similar visual attributes are placed closely.
Due to the shared weights, we believe this also indicates the good performance in tag estimator.

\section{Generative Adversarial Network}

\subsection{Vanilla GAN}
Generative Adversarial Networks proposed by Goodfellow et at.\cite{GAN} are implicit generative models. 
It proves to be an effective and efficient way to generate highly photo-realistic images in an unsupervised and likelihood-free manner\cite{DCGAN}. 
GAN uses a generator network $G$ to generate samples from $P_G$. This is done by transforming a latent noise variable $z \sim P_{noise}$ into a sample $G(z)$. The original GAN uses a min-max game strategy to train the generator $G$, imposing another network $D$ to distinguish samples from $G$ and real samples. Formally, the objective of GAN can be expressed as
\[
\min_G \max_D \mathcal{L}(D,G) = \mathbb{E}_{x \sim P_{data}}[\log D(x)] + \\  \mathbb{E}_{x \sim P_{noise}}[\log(1-D(G(z)))] \,.
\]
In this formula, the discriminator $D$ try to maximize the output confidence score from real samples. 
Meanwhile, it also minimizes the output confidence score from fake samples generated by $G$. 
On contrast, the aim of $G$ is to maximize the $D$ evaluated score for its outputs, which can be viewed as deceiving $D$.

\subsection{Improved training of GAN} \label{improved_gan_training}
Despite the impressive results of GAN, it is notoriously hard to train properly GAN. \cite{arjovsky2017towards} showed the 
$P_G$ and $ P_{data}$ may have non-overlap supports, 
so the Jensen-Shannon Divergence in the original GAN objective is constantly $0$, which leads to instability. \cite{arora2017generalization} argued that there may exist no equilibrium in the game between
generator and discriminator. 
One possible remedy is to use integral probability metric(IPM) based methods instead, e.g. Wasserstein distance\cite{arjovsky2017wasserstein}, Kernel MMD\cite{li2017mmd}, Cramer distance\cite{bellemare2017cramer}. 
Some recent GAN variants suggest using gradient penalty to stabilize GAN training\cite{gulrajani2017improved,bellemare2017cramer,roth2017stabilizing,kodali2017train}.
Mattya\cite{ChainerGANLib} compared several recent GAN variants under the same network architecture and measures their performance under the same metric.

Here, we use DRAGAN proposed by Kodali et al.\cite{kodali2017train} as the basic GAN model. 
As Mattya\cite{ChainerGANLib} shows, DRAGAN can give presumable results compare to other GANs, and it has the least computation cost among those GAN variants. Compare with Wasserstein GAN and its variants, DRAGAN can be trained under the simultaneous gradient descent setting, which make the training much faster.
In our experiments, we also find it is very stable under several network architectures, we successfully train the DRAGAN with a SRResNet\cite{ledig2016photo}-like generator, model details will be discussed in Section \ref{ganexperiments}.

The implementation of DRAGAN is thus straightforward: we only need to sample some points in local regions around real images and force those samples to have norm $1$ gradients with respect to the discriminator outputs. 
This can be done by adding a gradient penalty term to the generator loss. 
The flexibility of DRAGAN enables it to replace DCGAN in any GAN related tasks.

\subsection{GAN with labels}
Incorporating label information is important in our task
to provide user a way to control the generator.
Our utilization of the label information is inspired by ACGAN\cite{odena2016conditional}:
The generator $G$ receive random noise $z$ along with a 34-dimension vector $c$ indicate the corresponding attribute conditions. We add a multi-label classifier on the top of discriminator network, which try to predict the assigned tags for the input images. 

In detail, the loss is described as following:

\begin{align*}
\mathcal{L}_{adv}(D) &= - \mathbb{E}_{x \sim P_{data}}[\log D(x)] - \mathbb{E}_{x \sim P_{noise}, c \sim P_{cond}}[\log(1-D(G(z,c)))] \\
\mathcal{L}_{cls}(D) &= \mathbb{E}_{x \sim P_{data}}[\log P_D[label_x|x]] + \mathbb{E}_{x \sim P_{noise}, c \sim P_{cond}}[\log(P_D[c|G(z,c)])]  \\
\mathcal{L}_{gp}(D) &= \mathbb{E}_{\hat{x} \sim P_{perturbed_data}}[(||\nabla_{\hat{x}}D(\hat{x})||_2-1)^2 ]  \\
\mathcal{L}_{adv}(G) &= \mathbb{E}_{x \sim P_{noise}, c \sim P_{cond}}[\log(D(G(z,c)))]  \\
\mathcal{L}_{cls}(G) &= \mathbb{E}_{x \sim P_{noise}, c \sim P_{cond}}[\log(P_D[c|G(z,c)])]  \\
\mathcal{L}(D) &= \mathcal{L}_{cls}(D)  + \lambda_{adv}\mathcal{L}_{adv}(D)  +\lambda_{gp}\mathcal{L}_{gp}(D)  \\
\mathcal{L}(G) &=\lambda_{adv} \mathcal{L}_{adv}(G)  + \mathcal{L}_{cls}(G) \\
\end{align*}
where $P_{cond}$ indicates the prior distribution of assigned tags.  $\lambda_{adv}, \lambda_{gp}$ are balance factors for the adversarial loss and gradient penalty respectively.

\section{Experiments}
\subsection{Training details} \label{ganexperiments}
\subsubsection{Model architecture}
\begin{figure}
  \centering
  \includegraphics[width=5.5in]{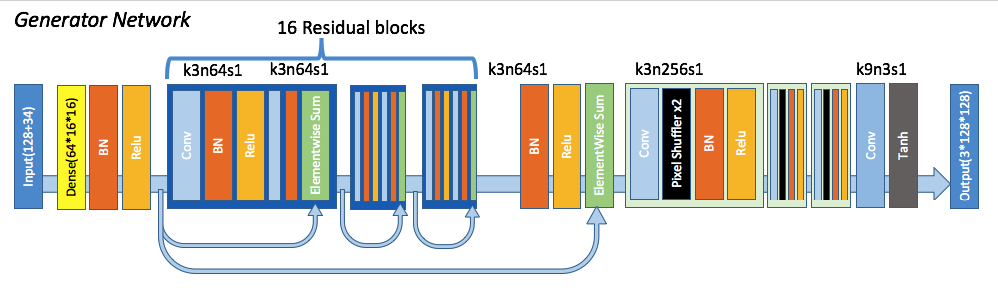}
  \caption{Generator Architecture}
  \label{fig:gen}
\end{figure}
\begin{figure}
  \centering
  \includegraphics[width=5.5in]{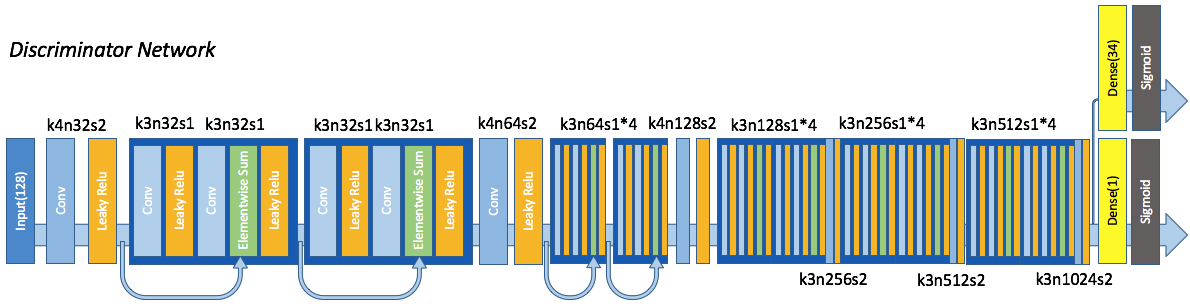}
  \caption{Discriminator Architecture}
  \label{fig:dis}
\end{figure}

The generator's architecture is shown in Figure \ref{fig:gen}, which is a modification from SRResNet\cite{ledig2016photo}. The model contains 16 ResBlocks and uses 3 sub-pixel CNN\cite{shi2016real} for feature map upscaling.
Figure \ref{fig:dis} shows the discriminator architecture, which contains 10 Resblocks in total. All batch normalization layers are removed in the discriminator, since it would bring correlations within the mini-batch, which is undesired for the computation of the gradient norm. 
We add an extra fully-connected layer to the last convolution layer as the attribute classifier. 
All weights are initialized from a Gaussian distribution with mean $0$ and standard deviation $0.02$.

\subsubsection{Hyperparameters}

We find that the model achieve best performance with $\lambda_{adv}$ equaling to the number of attributes, as Zhou et al.\cite{zhou2017generative} gives a detailed analysis of the gradient in the condition of ACGAN. 
Here, we set $\lambda_{adv}$ to $34$ and $\lambda_{gp}$ to $0.5$ in all experiments. 
All models are optimized using Adam optimizer\cite{kingma2014adam} with $\beta_1$ equaling $0.5$. 
We use a batch size of 64 in the training procedure. 
The learning rate is initialized to $0.0002$ and exponentially decease after $50000$ iterations of training.

\subsubsection{Model Training}
\begin{figure}[h]
  \centering
  \includegraphics[width=\textwidth]{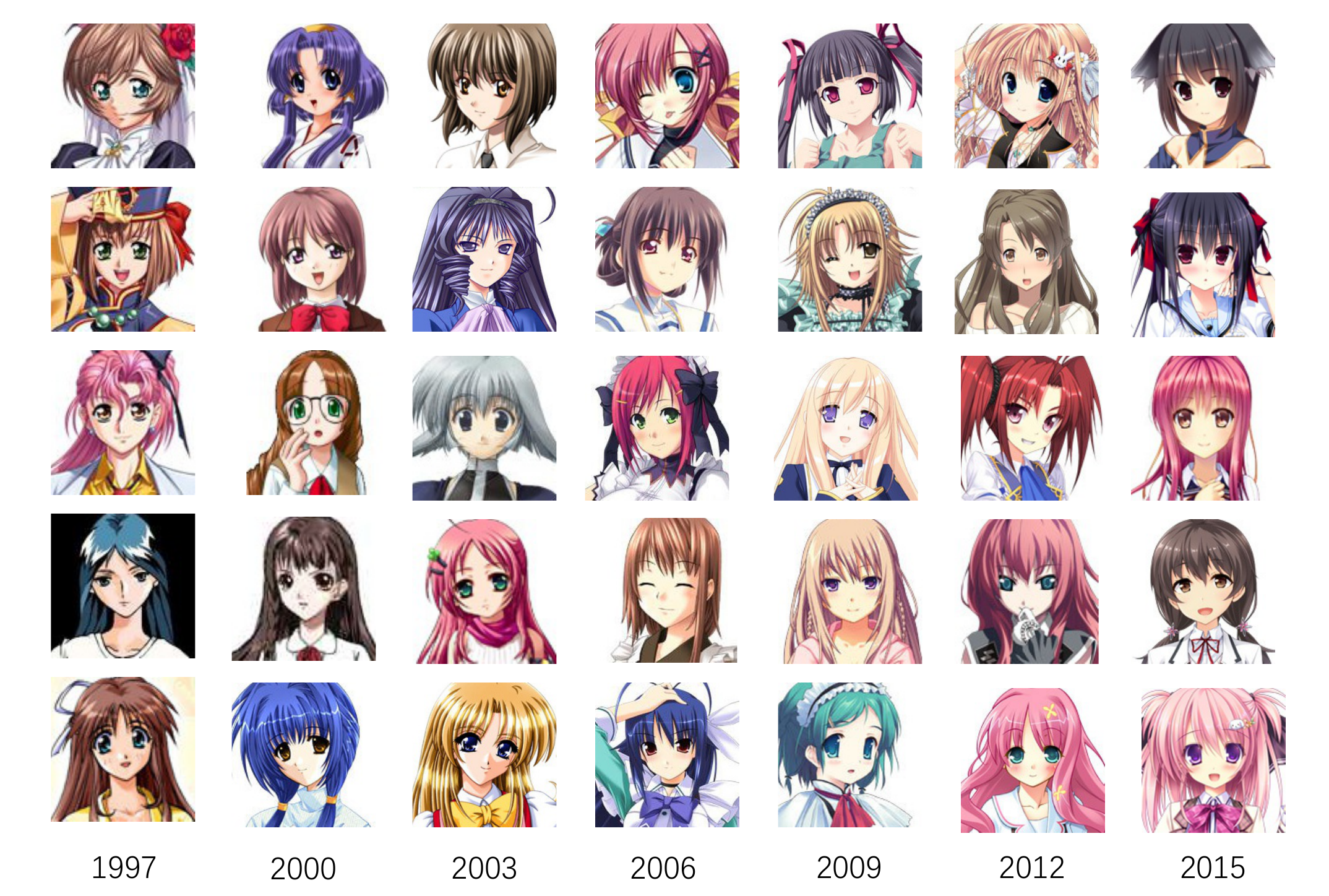}
  \caption{Sample images in different release years.}
  \label{fig:chara_year}
\end{figure}

\begin{figure}[h]
  \centering
  \includegraphics[width=\textwidth]{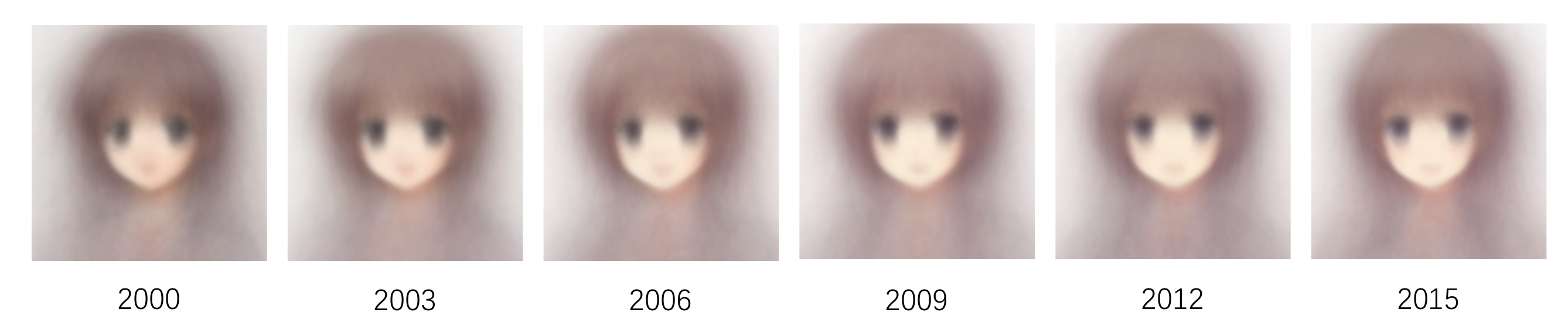}
  \caption{Average images from 1000 samples in different release years.}
  \label{fig:avg_face}
\end{figure}
The technology of making game characters and CGs is evolving continuously, therefore the release year of the game plays an important role for the visual aspect of image quality. 
As we can see in the Figure \ref{fig:chara_year}, characters before 2003 look old-fashioned, while characters in the recent games is cuter and have better visual quality. Appendix \ref{image_distribution} shows the distribution of images in our dataset.

We train our GAN model using only images from games released after 2005
and with scaling all training images to a resolution of 128*128 pixels. 
This gives 31255 training images in total.

On the conditional generation of images, the prior distribution of labels $P_{cond}$ is critical, 
especially when labels are not evenly distributed. 
In our case, there are only $49$ training images assigned with the attribute ``orange eyes'' while $8861$ images are assigned with the attribute ``blue eyes''.

But we don't take this in to account in the training stage.
To sample related attributes for the noise, we use the following strategy. 
For the hair and the eye color, we randomly select one possible color with uniform distribution. For other attributes, we set each label independently with a probability of 0.25. 
\subsection{Generated Results}
\begin{figure}[h]
  \centering
  \includegraphics[width=\textwidth]{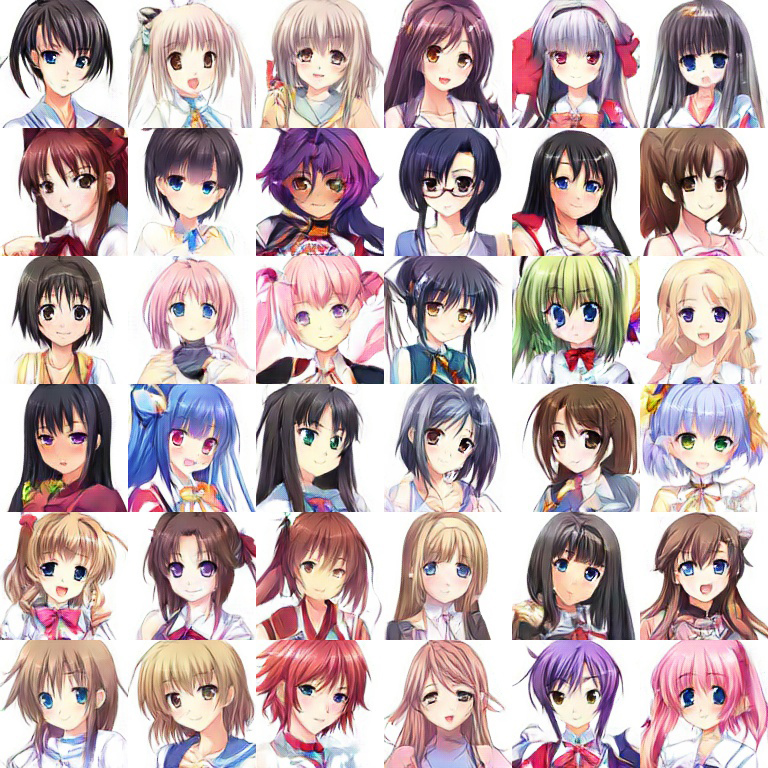}
  \caption{Generated samples}
  \label{fig:samples}
\end{figure}

Figure \ref{fig:samples} shows images generated from our model. 
Different from the training stage, we set the probability of each label based on the corresponding empirical distribution in the training dataset here.

\begin{figure}[h]
  \centering
  \includegraphics[width=\textwidth]{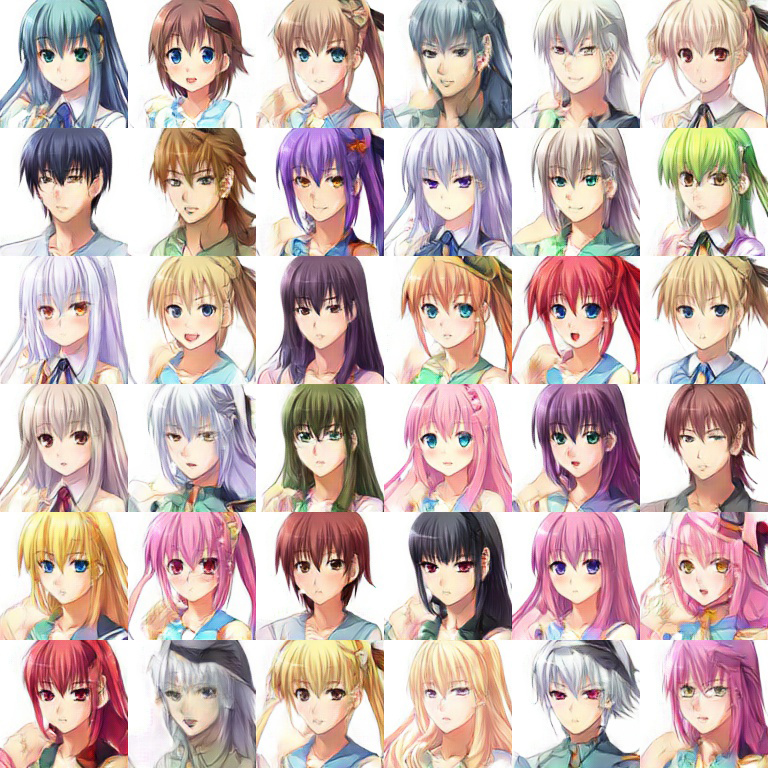}
  \caption{Generated images with fixed noise part and random attributes}
  \label{fig:fix_noise}
\end{figure}

By fixing the random noise part and sampling random conditions, the model can generate images have similar major visual features (e.g. face shapes, face directions). 
Figure \ref{fig:fix_noise} is an example of that. It is also an evidence of the generalization ability of visual concepts learnt from corresponding labels, indicating that our model can avoids the brute-force memorization of training samples. 

Another phenomenon we empirically observed is that the random  noise part heavily inference the quality of the final result. Some noise vector can give good samples no matter what conditioned on, while some other noise vectors are easier to produce distorted images.

\begin{figure}[h]
  \centering
  \includegraphics[width=\textwidth]{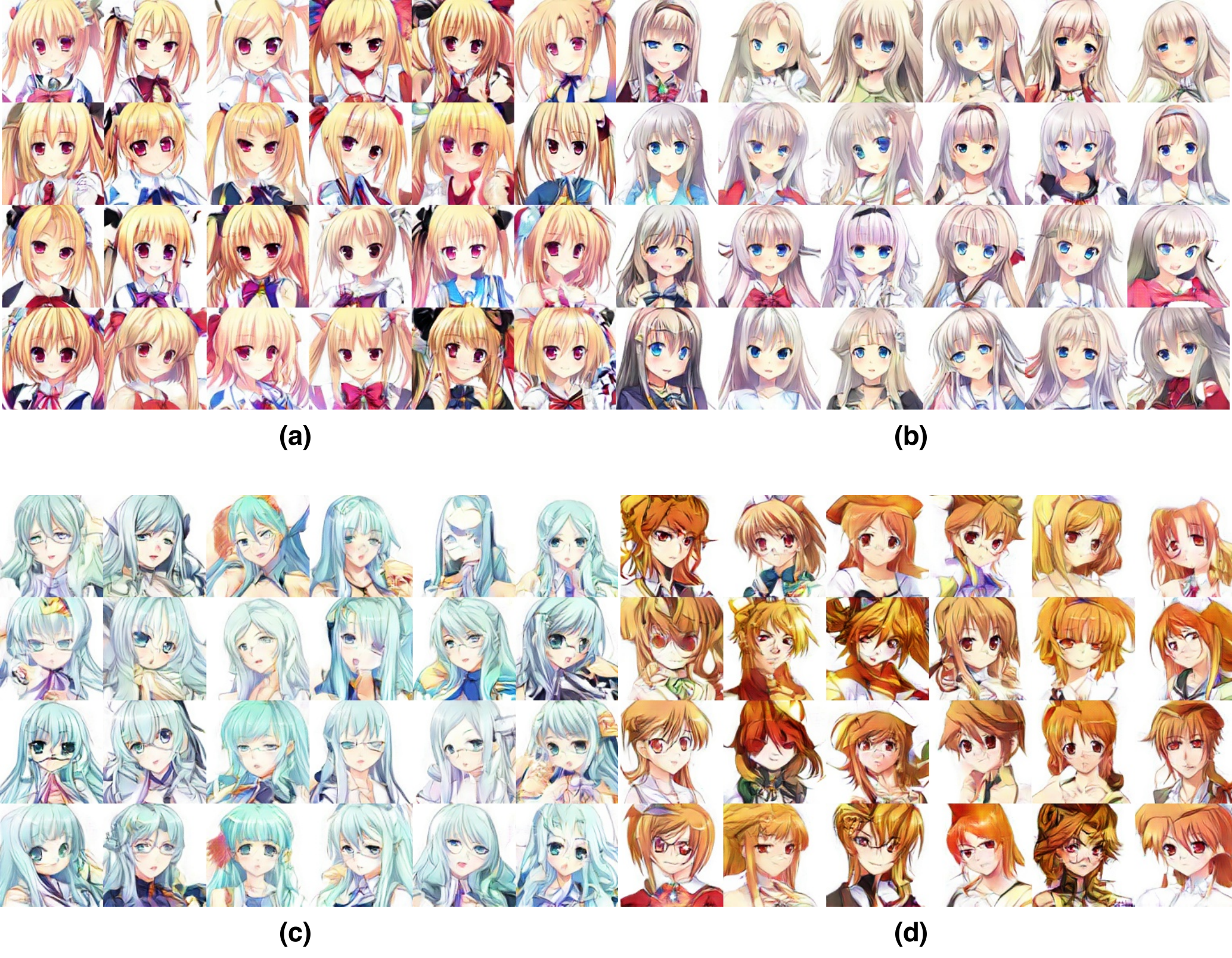}
  \caption{Generated images under fixed conditions. (a) blonde hair, twintails, blush, smile, ribbon, red eyes (b) silver hair, long hair, blush, smile, open mouth, blue eyes (c) aqua hair, long hair, drill hair, open mouth, glasses, aqua eyes (d) orange hair, ponytail, hat, glasses, red eyes, orange eyes}
  \label{fig:fix_attr}
\end{figure}

As Table \ref{tab:tag_count} states, labels are not evenly distributed in our training dataset, which results that some combinations of attributes cannot give good images.  In Figure \ref{fig:fix_attr}, (a)(b) are generated with well learned attributes like ``blonde hair'', ``blue eyes''.  On contrast, (c)(d) are associated with ``glasses'', ``drill hair'', which is not well learned because of the insufficiency of corresponding training images. All characters in (a)(b) appear to be attractive, but most characters in (c)(d) are distorted.

\subsection{Quantitative Analysis}
\subsubsection{Attribute Precision}

\begin{table}[h]
  \small
  \begin{tabular}{c c c c c c c}
    \toprule
    blonde hair & brown hair & black hair & blue hair & pink hair & purple hair & green hair \\
    1.00 &   1.00&   1.00 &   0.70 &   0.80 &   0.75&    0.90  \\ \midrule
    red hair & silver hair & white hair & orange hair & aqua hair & gray hair & long hair \\ 
    0.95 &    0.85&    0.60&    0.65&    1.00&     0.35&  1.00
  \\  \midrule
    short hair & twintails & drill hair & ponytail & blush & smile & open mouth \\ 
    1.00 &   0.60&   0.20&   0.45&   1.00&   0.95&   0.95 \\
    \midrule
    hat & ribbon & glasses & blue eyes & red eyes & brown eyes & green eyes \\
     0.15 &   0.85 &  0.45&   1.00&   1.00&   1.00&   1.00 \\ \midrule
    purple eyes & yellow eyes & pink eyes & aqua eyes & black eyes & orange eyes \\
    0.95 &   1.00 &    0.60 &    1.00&    0.80&    0.85 \\ \bottomrule \\
  \end{tabular}
  \caption{Precision of each label}
  \label{tab:label_precision}
\end{table}

To evaluate how each tag affect the output result, we measure the precision of the output result when the certain label is assigned. With each target, we fix the target label to true, and sample other labels in random. For each label, 20 images are drawn from the generator. Then we manually check generated results and judge whether output images behave the fixed attribute we assigned. Table \ref{tab:label_precision} shows the evaluation result. From the table we can see that  
compared with shape attributes(e.g. ``hat'', ``glasses''), 
color attributes are easier to learn. 
Notice that the boundary between similar colors like ``white hair'', ``silver hair'', ``gray hair '' is not clear enough,. Sometimes people may have troubles to classify those confusing colors. This phenomenon lead to low precision scores for those attributes in our test. 

Surprisingly, some rare color attributes like ``orange eyes'', ``aqua hair'', ``aqua eyes'' have a relative high precisions even though samples containing those attributes are less than 1\% in the training dataset.  We believe visual concepts related to colors are simple enough for the generator to get well learned with a extremely small number of training samples.

On contrast, complex attribute like ``hat'', ``glasses'', ``drill hair'' are worst behaved attributes in our experiments. When conditioned on those labels, generated images are often distorted and difficult to identify. 
Although there are about 5\% training samples assigned with those attributes, the complicated visual concept they implied are far more accessible for the generator to get well learned.

\subsubsection{FID Evaluation}
One possible quantitative evaluation method for GAN model is Fr\'{e}chet Inception Distance(FID) proposed by Heusel et al.\cite{heusel2017gans}. To calculate the FID, they use a pre-trained CNN(Inception model) to extract vision-relevant features from both real and fake samples. The real feature distribution and the fake feature distribution  are approximated with two guassian distributions. Then, they calculate The Fr\'{e}chet distance(Wasserstein-2 distance) between two guassian distributions and serve the results as a measurement of the model quality.

The Inception model trained on ImageNet is not suitable for extracting features of anime-style illustrations, since there is no such images in the original training dataset.
Here, we replace the model with Illustration2vec feature extractor model for better measurement of visual similarities between generated images and real images.

\begin{table}[h]
  \centering
  \begin{tabular}{ccc}
  \toprule
  Model & Average FID & MaxFID-MinFID \\
  \midrule
  DCGAN Generator+DRAGAN &5974.96 & 85.63 \\
  Our Model & 4607.56 & 122.96  \\
  \bottomrule \\
  \end{tabular}
  \caption{FID of our model and baseline model}
  \label{tab:fid_models }
 \end{table}
To evaluate the FID score for our model, we sample 12800 images from real dataset, then generate a fake sample by using the corresponding conditions for each samples real images.
After that we feed all images to the Illustation2vec feature extractor and get a 4096-dimension feature vector for each image. FID is calculated between the collection of feature vectors from real samples and that from fake samples.

For each model, we repeat this process for 5 times and measure the average score of 5 FID calculation trails. Table \ref{tab:fid_models } shows the result comparing our model with the baseline model.  We observe that SRResNet based model can achieve better FID performance evenly with less weight parameters. 

\begin{figure}[h]
  \centering
  \includegraphics[width=3.5in]{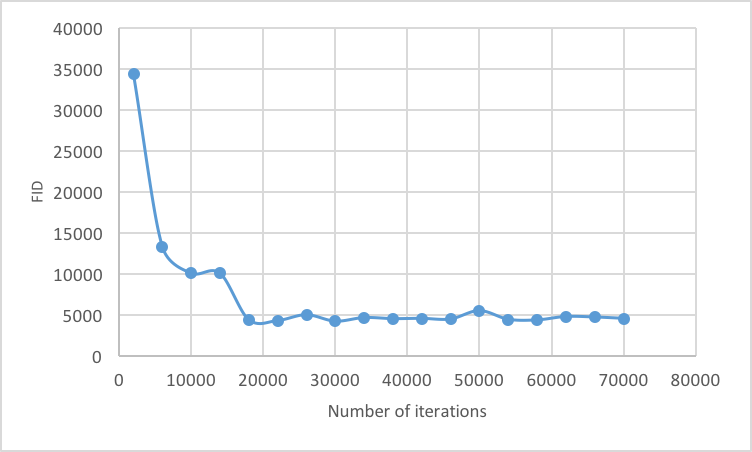}
  \caption{FID decrease and converge to a certain value during the training procedure}
  \label{fig:fid_iterations}
\end{figure}

\subsection{Website Interface}
In order to make our model more accessible, we build a website interface\footnote{\url{http://make.girls.moe}} for open access. We impose WebDNN\footnote{\url{https://mil-tokyo.github.io/webdnn/}} and convert the trained Chainer model to the WebAssembly based Javascript model. The web application is built with React.js.  

Keeping the size of generator model small would be a great benefit when hosting a web browser based deep learning service. This is because user are required to download the model before the computation every time, bigger model results much more downloading time which will affect the user experience.
Replacing the DCGAN generator by SRResNet generator can make the model 4x smaller, so the model downloading time can be reduced by a large margin. 

\begin{figure}[h]
  \centering
  \includegraphics[width=\textwidth]{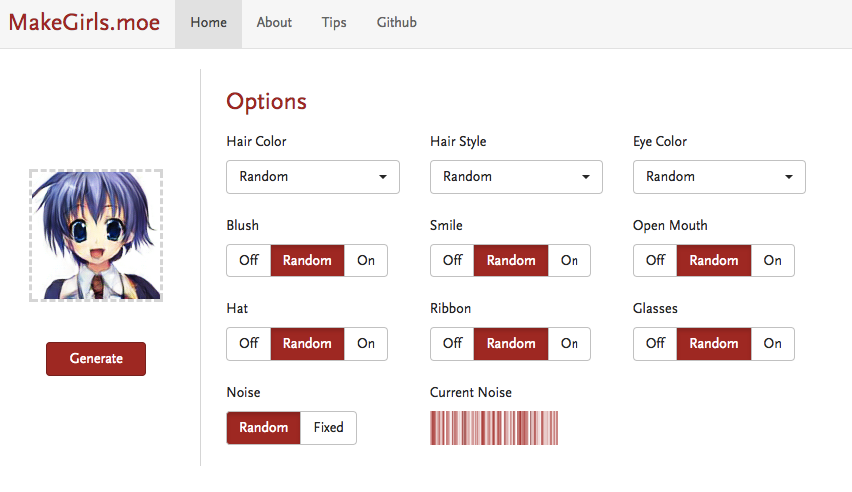}
  \caption{Our website}
  \label{fig:website}
\end{figure}

Users can manually assign attributes to the generator, and all unassigned attributes will be sampled based on empirical distribution of the training dataset. All computations are done on the client side. It takes about $6\sim 7$s to generate one images on average, the detailed performance test is discussed in Appendix \ref{inference_time}.

\subsection{Super-Resolution}
As Appendix \ref{image_resolution} shows, 
the resolution of available training images are not high, 
making generation of high resolution facial images from the GAN model directly is difficult. 
Here we try to build another Super-Resolution network 
specifically for generation extra high resolution animation style images to overcome the limitation. 

\begin{figure}[h]
  \centering
  \includegraphics[width=3.5in]{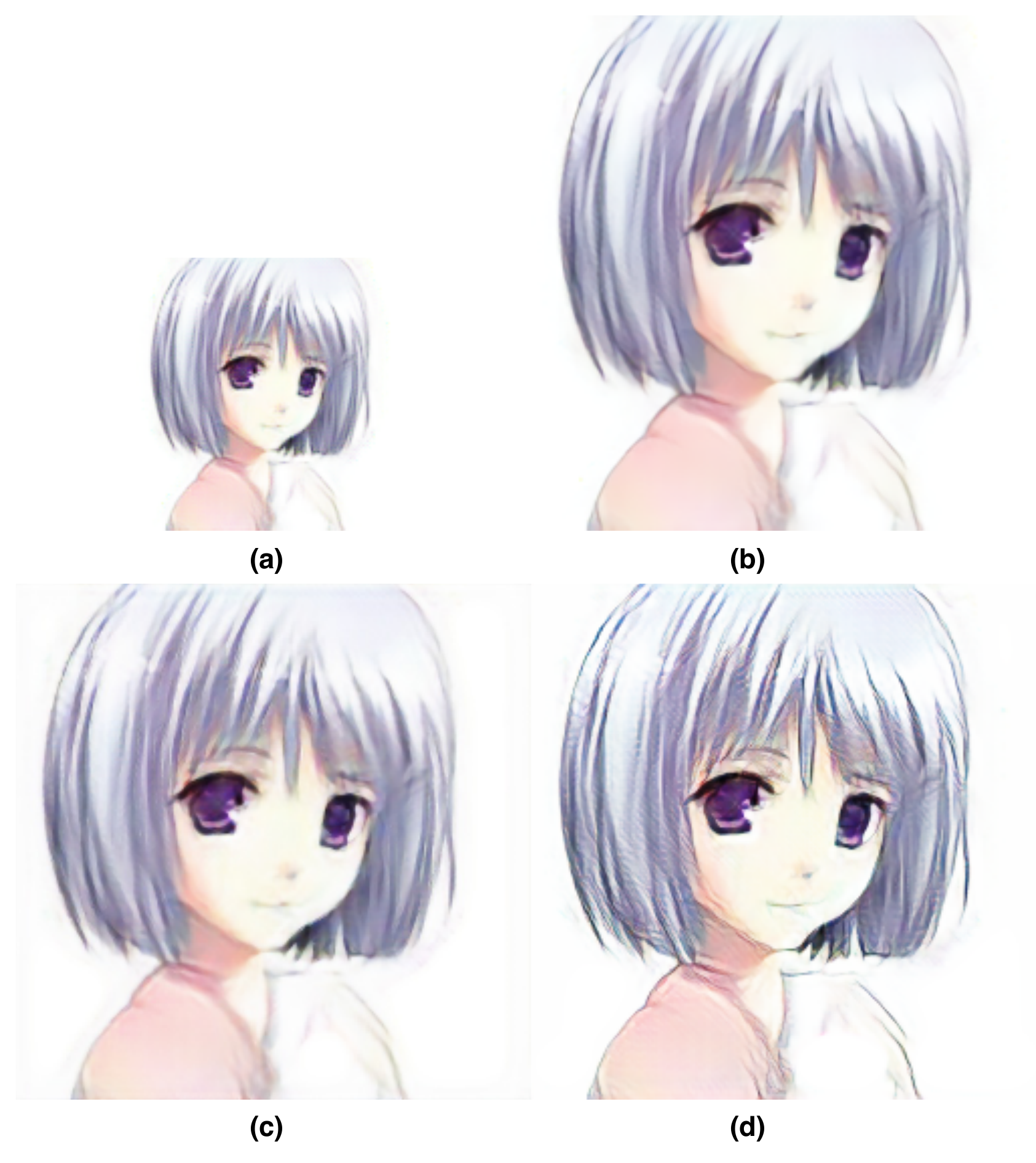}
  \caption{Results of 2x super-resolution. (a) The original image. (b) Result from waifu2x (c)SRGAN with lower adversarial loss (d) SRGAN with higher adversarial loss}
  \label{fig:sr}
\end{figure}

Image (b) in Figure \ref{fig:sr} shows the 2x upscaled image with waifu2x\footnote{\url{http://waifu2x.udp.jp/index.ja.html}}, which is blurred. 
\cite{ledig2016photo} gives a comparison of GAN-based super-resolution model and traditional MSE-based super-resolution model. 
Their result shows that GAN based models can bring more high-frequency details to the upscaled images than MSE only models. 
This is preferred in our situations, so we choose to implement SRGAN\cite{ledig2016photo}  as our super-resolution model. 

Image (c) (d) show our attempts of training a SRGAN. (c) is trained with a low adversarial loss weight and (d) is trained with high adversarial loss weight. 
We observe that as the weight of the adversarial loss increasing, 
the upscaled image looks sharper.
However, this would bring more undesired artifacts to output images.

We observe that the visual quality of anime-style images are more sensitive to extra artifacts than real photos. 
We hypothesize that it is due to the fact that color/texture patterns in anime-style images are much simpler and clearer than real photos, 
any artifacts would largely damage the color/texture patterns and make the result looks dirty and messy. 
This obstacle limit the usage of GAN in our super-resolution networks.

Discouragingly, we failed to find a model balanced well between the sharpness level and the artifact strengths, 
so we choose not add the super-resolution model to our website for now,
and leave the exploration of anime image focused super resolution for future work.

\section{Conclusion}

We explore the automatic creation of the anime characters in this work. 
By combining a clean dataset and several practicable GAN training strategies, 
we successfully build a model which can generate realistic facial images of anime characters. 
We also make available an easy-to-use website service online.

There still remain some issues for us for further investigations. 
One direction is how to improve the GAN model when class labels in the training data are not evenly distributed. 
Also, quantitative evaluating methods under this scenario should be analyzed, 
as FID only gives measurement when the prior distribution of sampled labels equals to the empirical labels distribution in the training dataset.
This would lead to a measure bias when labels in the training dataset are unbalanced.

Another direction is to improve the final resolution of generated images. 
Super-resolution seems a reasonable strategy, 
but the model need to be more carefully designed and tested. 

We hope our work would stimulate more studies on generative modeling of anime-style images and eventually 
help both amateurs and professionals design and create new anime characters.

\section{Acknowledgement}

This paper is presented as a Doujinshi in Comiket 92, summer 2017, with the booth number 05a, East-U, Third Day. 

The work is done when Yanghua Jin works as a part-time engineer in Preferred Networks, Japan. 
Special thanks to Eiichi Matsumoto, Taizan Yonetsuji, Saito Masaki, Kosuke Nakago from Preferred Networks for insightful directions and discussions. 

The cover illustration of Doujinshi is created by Zhihao Fang, and Jiakai Zhang helps create the website.

\small
\bibliographystyle{plain}
\bibliography{ref.bib}
\newpage
\section{Appendix}
\subsection{SQL query on ErogameScape} \label{SQL code}

\begin{lstlisting}[language=SQL]
SELECT g.id, g.gamename, g.sellday, 
      'www.getchu.com/soft.phtml?id=' || g.comike as links
FROM gamelist g
WHERE g.comike is NOT NULL
ORDER BY g.sellday
\end{lstlisting}

\subsection{Dataset images distribution} \label{image_distribution}
\begin{figure}[h]
  \centering
  \includegraphics[width=\textwidth]{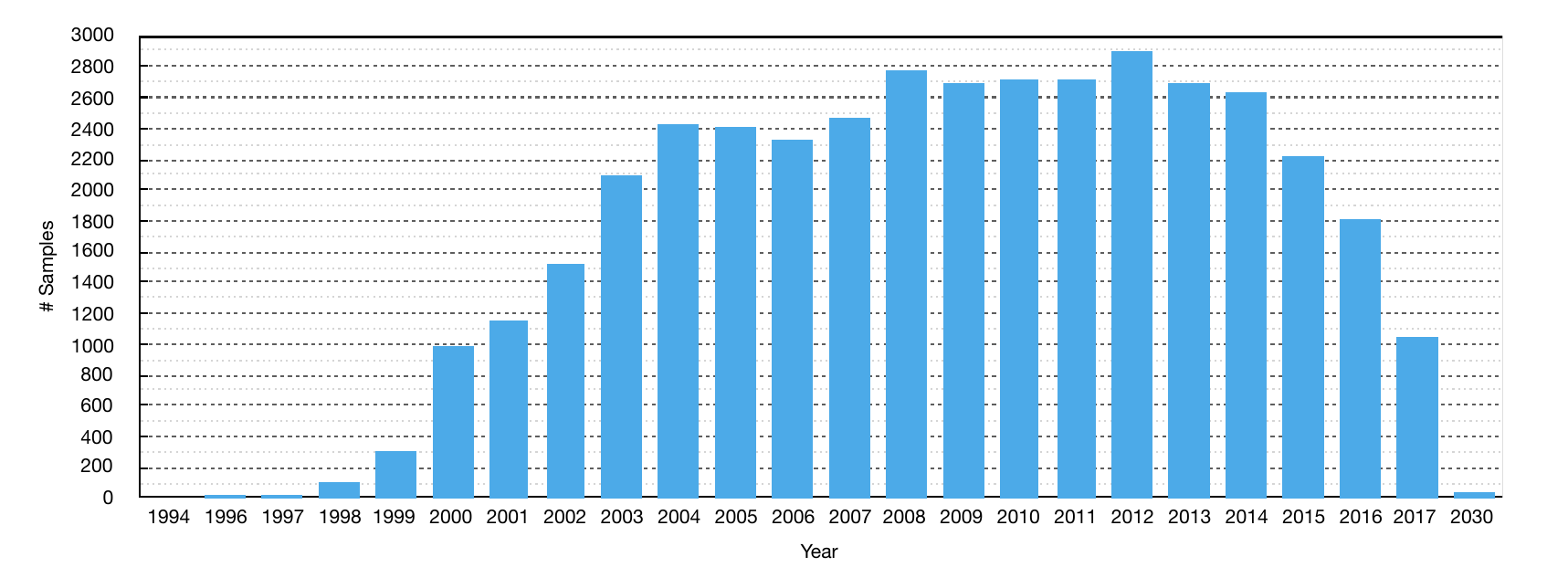}
  \caption{Available dataset images by release years. Note that year=2030 means release year undetermined}
  \label{fig:game_year}
\end{figure}

\label{image_resolution}
\begin{figure}[h]
  \centering
  \includegraphics[width=\textwidth]{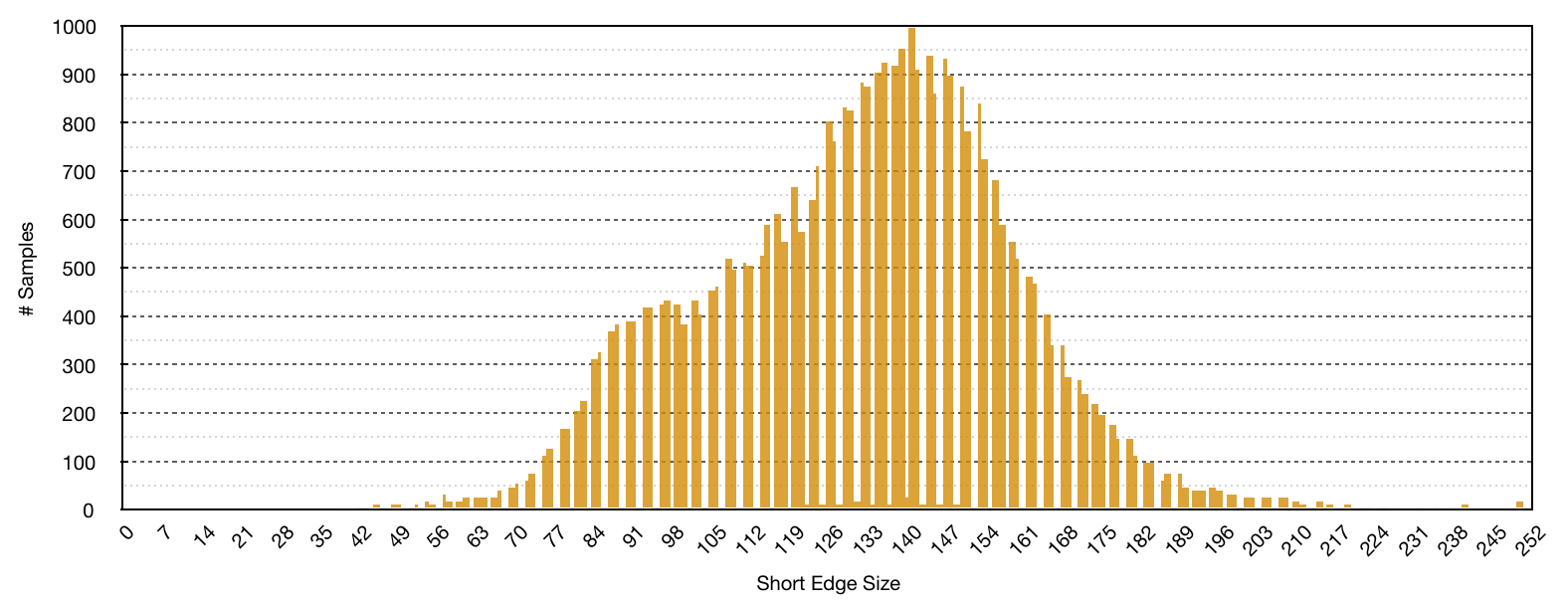}
  \caption{Available dataset images by short edge size.}
  \label{fig:img_size}
\end{figure}
\newpage
\subsection{Interpolation between random generated samples  } 
\begin{figure}[h]
  \centering
  \includegraphics[width=\textwidth]{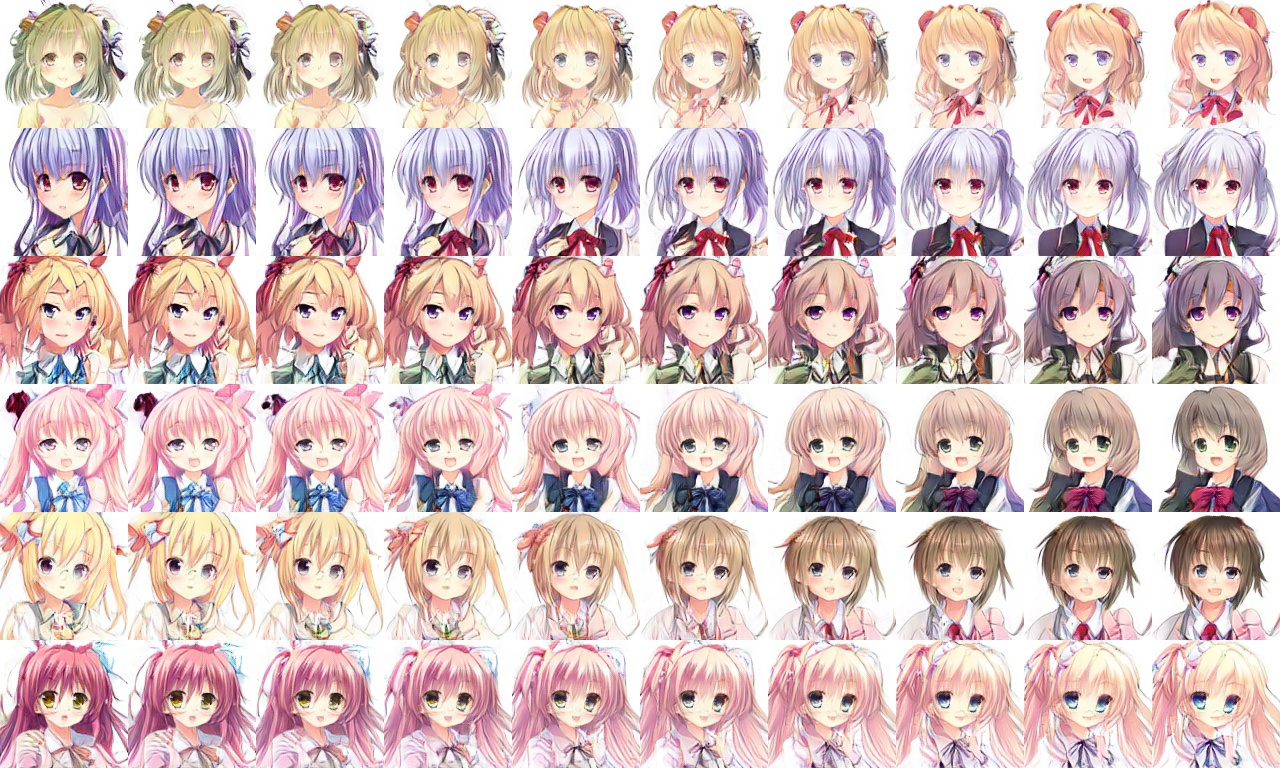}
  \caption{Samples in the first column and the last columns are randomly generated under different combinations of conditions. Although label controlling variables are assigned with discrete values in the training stage, the result shows that those discrete attributes are still meaningful under the continuous setting.}
  \label{fig:interpolation}
\end{figure}

\subsection{Approximate inference time} \label{inference_time}
\begin{table}[h]
  \centering
  \begin{tabular}{c  c  c  c}
  \toprule
  Processor & Operation System & Web Browser & Execution Time (s) \\ 
  \midrule
  I7-6700HQ & macOS Sierra & Chrome 59.0 & 5.55 \\
  I7-6700HQ & macOS Sierra & Safari 10.1 & 5.60 \\
  I5-5250U & macOS Sierra & Chrome 60.0 & 7.86 \\
  I5-5250U & macOS Sierra & Safari 10.1 & 8.68 \\
  I5-5250U & macOS Sierra & Safari Technology Preview 33*  & <0.10 \\
  I5-5250U & macOS Sierra & Firefox 34 & 6.01 \\
  I3-3320 & Ubuntu 16.04 & Chromium 59.0 & 53.61 \\
  I3-3320 & Ubuntu 16.04 & Firefox 54.0 & 4.36 \\
  iPhone 7 Plus & iOS 10 &  Chrome & 4.82 \\
  iPhone 7 Plus & iOS 10 &  Safari & 3.33 \\
  iPhone 6s Plus & iOS 10 & Chrome & 6.47 \\
  iPhone 6s Plus & iOS 10 & Safari & 6.23 \\
  iPhone 6 Plus & iOS 10 & Safari & 11.55 \\
  \bottomrule
  \end{tabular}
  \caption{Approximate inference time on several different environments, note that for Safari Technology Preview, the computation is done by WebGPU ,while for other browsers, the computation is done by WebAssembly. We can see that firefox is better optimized with WebAssembly and faster than other browsers }
  \label{tab:model_executing_time }
 \end{table}

\end{document}